\pdfoutput=1
\documentclass[11pt]{article}

\usepackage[preprint]{acl}

\usepackage{times}
\usepackage{latexsym}

\usepackage[T1]{fontenc}
\usepackage[utf8]{inputenc}

\usepackage{microtype}

\usepackage{inconsolata}

\usepackage{microtype}
\usepackage{graphicx}
\usepackage{booktabs}
\usepackage{float}
\usepackage{subfig}

\usepackage{tikz}
\usepackage{pgfplots}
\usepackage{pgfplotstable}

\usepackage{amsmath}
\usepackage{amssymb}
\usepackage{mathtools}
\usepackage{amsthm}
\renewcommand{\vec}[1]{\boldsymbol{#1}}

\usepackage[capitalize,noabbrev]{cleveref}
\theoremstyle{plain}

\theoremstyle{definition}

\theoremstyle{remark}

\newcommand{\minisection}[1]{\noindent{\textbf{#1}}.}
\usepackage{multirow}
\usepackage{tabularx}
\usepackage{bbm}
\usepackage{makecell}
\usepackage{enumitem}

\usepackage{moresize}

\usepackage{threeparttable}

\usepackage{pifont}
\usepackage{fontawesome}

\title{Efficient OpAmp Adaptation for Zoom Attention to Golden Contexts}

\author{
\textbf{Haoyuan Wu\textsuperscript{$\spadesuit$}\thanks{These authors contributed equally to this work.}},
\textbf{Rui Ming\textsuperscript{$\spadesuit$}$^*$},
\textbf{Haisheng Zheng\textsuperscript{$\heartsuit$}},
\textbf{Zhuolun He\textsuperscript{$\spadesuit$,$\clubsuit$}},
\textbf{Bei Yu\textsuperscript{$\spadesuit$}}
\\\\
\textsuperscript{$\spadesuit$}The Chinese University of Hong Kong, Hong Kong SAR \\
\textsuperscript{$\heartsuit$}Shanghai Artificial Intelligent Laboratory, China \\
\textsuperscript{$\clubsuit$}ChatEDA Tech, China\\
\texttt{\{hywu24,byu\}@cse.cuhk.edu.hk} \\
}

\begin{document}
\maketitle

\begin{abstract}
Large language models (LLMs) have shown significant promise in question-answering (QA) tasks, particularly in retrieval-augmented generation (RAG) scenarios and long-context applications. 
However, their performance is hindered by noisy reference documents, which often distract from essential information. 
Despite fine-tuning efforts, Transformer-based architectures struggle to prioritize relevant content. 
This is evidenced by their tendency to allocate disproportionate attention to irrelevant or later-positioned documents. 
Recent work proposes the differential attention mechanism to address this issue, but this mechanism is limited by an unsuitable common-mode rejection ratio (CMRR) and high computational costs. 
Inspired by the operational amplifier (OpAmp), we propose the OpAmp adaptation to address these challenges, which is implemented with adapters efficiently. 
By integrating the adapter into pre-trained Transformer blocks, our approach enhances focus on the golden context without costly training from scratch. 
Empirical evaluations on noisy-context benchmarks reveal that our Qwen2.5-OpAmp-72B model, trained with our OpAmp adaptation, surpasses the performance of state-of-the-art LLMs, including DeepSeek-V3 and GPT-4o.
\end{abstract}

\section{Introduction}

Recent advancements in large language models (LLMs)~\cite{openai2023gpt4, dubey2024llama3,yang2024qwen2_5,liu2024deepseekv3} have demonstrated remarkable capabilities in understanding, generating, and reasoning across diverse domains, significantly advancing their application in various fields. 
Among these applications, question answering (QA) based on provided contexts has emerged as one of the most prominent use cases for LLMs.

As LLMs' capabilities continue to evolve and user expectations grow, users increasingly supply multiple documents retrieved in Retrieval-Augmented Generation (RAG) scenarios or long-context reference documents to guide LLMs in generating contextually relevant responses.
However, in practice, such retrieved documents or long-context references often contain substantial noise, including information irrelevant to the user's query.
Recent studies~\cite{ye2025difftrans,liu2024lost} highlight a critical challenge that LLMs frequently struggle to accurately identify and extract key information from these noisy contexts, limiting their effectiveness in real-world applications.

\begin{figure}[!tb]
    \centering
    \includegraphics[width=0.928\linewidth]{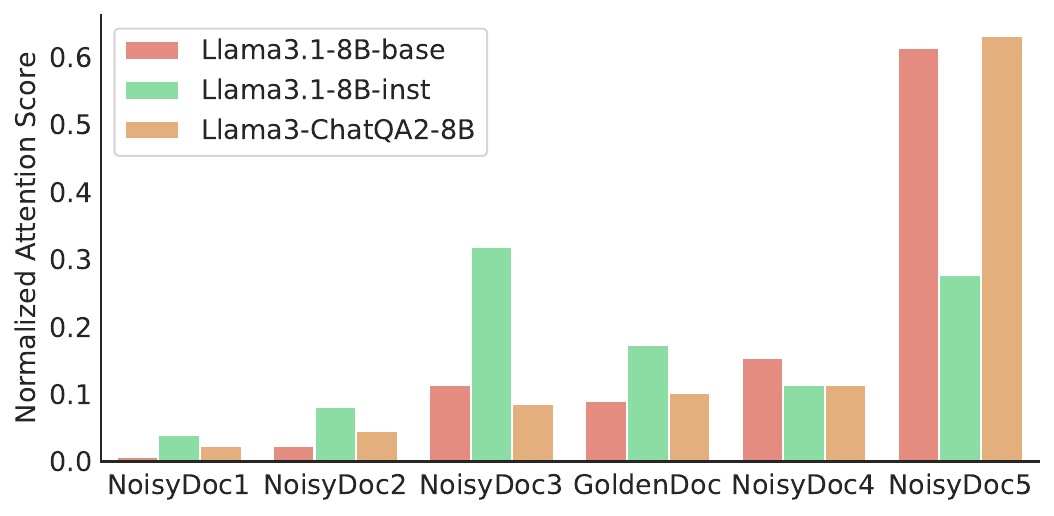} 
    \caption{Normalized attention score. Transformers often miss the golden document in a noisy context.} 
    \label{fig:motivation}
\end{figure}

As illustrated in \Cref{fig:motivation}, we visualize the normalized attention scores assigned to retrieved documents in the RAG scenario, which includes various noisy documents and a single golden document. 
The task involves identifying the correct answer within noisy contexts. 
Our analysis evaluates several LLMs, including Llama3.1-8B-base~\cite{meta2024llama3_1}, Llama3.1-8B-inst~\cite{meta2024llama3_1}, and Llama3-ChatQA2-8B~\cite{xu2024chatqa2}, the latter of which has been fine-tuned specifically for long-context and RAG applications. 
The visualization demonstrates that the Transformer architecture tends to allocate only a small proportion of attention scores to the golden document, while disproportionately focusing on irrelevant or later-positioned documents. 
% Notably, ChatQA2, despite its fine-tuning for long-context and RAG tasks, tends to over-attend to documents positioned later in the sequence rather than the golden document. 
% Similarly, the aligned LLM struggles to maintain focus on relevant information in noisy environments.
These findings highlight a persistent challenge for Transformer-based architectures including effectively identifying and prioritizing relevant documents in the presence of noise. 
The issue~\cite{ye2025difftrans} arises from the non-negligible allocation of attention scores to irrelevant content, which ultimately obscures the correct answer and undermines model performance.

\begin{figure}[tb!]
    \centering
    \includegraphics[width=0.928\linewidth]{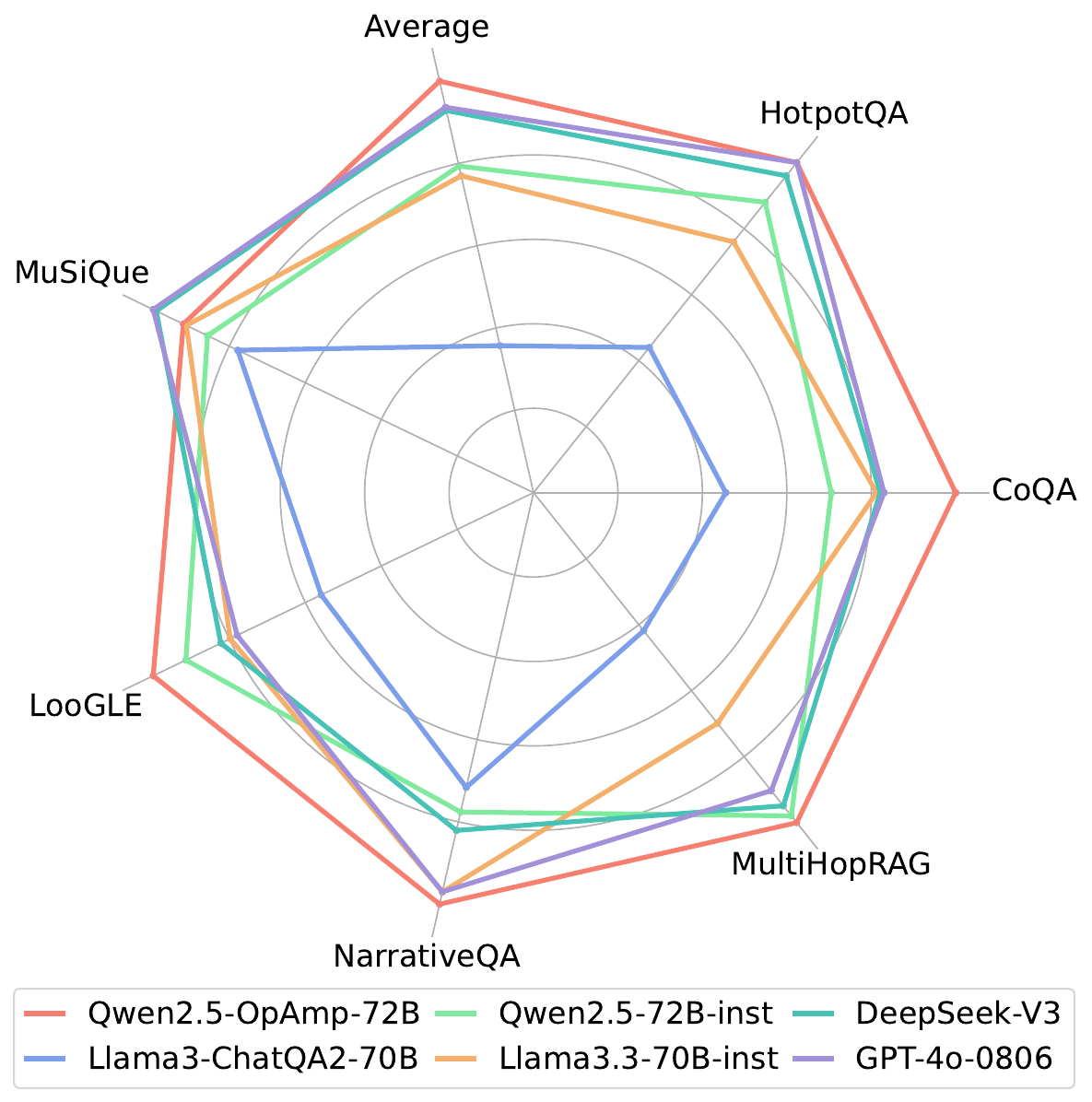} 
    \caption{Qwen2.5-OpAmp-72B achieves the best average performance in various noisy-context benchmarks compared to current SOTA LLMs.}
    \label{fig:radar}
\end{figure}

\citet{ye2025difftrans} propose a differential attention mechanism designed to mitigate attention noise through differential denoising, inspired by the principles of differential amplifiers in electrical engineering. 
However, differential amplifiers are effective in scenarios requiring a high common-mode rejection ratio (CMRR) considering that they only focus on differential gain. 
This is unsuitable for attention denoising in the Transformer block.
Training a differential transformer from scratch entails great computation costs and introduces significant risks, further limiting its practical applicability.

Inspired by the operational amplifiers (OpAmp), we introduce OpAmp adaptation with adapters, an efficient approach for refining the attention mechanism to enhance focus on the most relevant context leveraging parameter-efficient fine-tuning (PEFT) techniques. 
The OpAmp adaptation enables simultaneous control of differential gain and common-mode gain through the management of the CMRR. 
Building on the OpAmp design, our approach facilitates the training of OpAmp models using pre-trained Transformer architectures, eliminating the need for training from scratch. 
This strategy significantly reduces computational costs compared to previous methods. 
As demonstrated in \Cref{fig:radar}, our Qwen2.5-OpAmp-72B model, trained with the OpAmp adaptation, achieves superior average performance across various noisy-context benchmarks compared to current state-of-the-art (SOTA) LLMs. 
Our contributions are as follows:
\begin{itemize}[itemsep=0pt,topsep=0pt,parsep=0pt]
    \item We introduce the OpAmp adaptation for zoom attention to the most relevant context in noisy contexts;
    \item Implement OpAmp adaptation with adapters, which are fine-tuned with our noisy context dataset, achieving significant improvements;
    \item Develop OpAmp models with our OpAmp adaptation method, surpassing current SOTA LLMs in various noisy-context benchmarks.
\end{itemize}
\section{Methods}

\subsection{Preliminaries}

\minisection{Adapters}
\citet{houlsby2019adapter} introduced the concept of integrating adapters into pre-trained transformer-based models for PEFT. 
This approach only fine-tunes the parameters introduced by the adapters while maintaining the pre-trained weights with large parameters unchanged.
An adapter module comprises two trainable matrices, $\vec{W}_1 \in \mathbb{R}^{d_1 \times d_2}$ and $\vec{W}_2 \in \mathbb{R}^{d_2 \times d_1}$, along with a non-linear activation function  $\phi(\cdot)$. 
Here, $d_1$ represents the feature dimension of the pre-trained weights, while $d_2$ denotes the hidden dimension of the inserted adapter, typically satisfying $d_2 \ll d_1$.
Given an input feature $\vec{H} \in \mathbb{R}^{N \times d_1}$, the output of the adapter module is expressed as:
\begin{equation}
    \vec{H}' = \phi(\vec{H}\vec{W}_1)\vec{W}_2 + \vec{H}.
\label{eq:adapter}
\end{equation}

\minisection{Attention}  
The self-attention mechanism~\cite{vaswani2017attention} serves as the foundational building block for LLMs~\cite{openai2023gpt4, dubey2024llama3, yang2024qwen2_5, liu2024deepseekv3}. 
Given a query feature $\vec{Q} \in \mathbb{R}^{N \times d}$, a key feature $\vec{K} \in \mathbb{R}^{N \times d}$, and a value feature $\vec{V} \in \mathbb{R}^{N \times d}$, the attention mechanism is computed as follows:  
\begin{align}
    \mathrm{Attn}(\vec{Q}, \vec{K}, \vec{V}) &= \vec{M}\vec{V}, \nonumber\\
    \vec{M} &= \mathrm{Softmax}\left(\frac{\vec{Q}\vec{K}^{\top}}{\sqrt{d}}\right),
\label{eq:attn}
\end{align}
where $N$ represents the number of tokens and $d$ denotes the dimensionality of the query, key, and value features.

\minisection{Differential Amplifier}  
The differential amplifier~\cite{sansen2007analog} is an electronic device designed to amplify the voltage difference between its two input signals while rejecting any voltage common to both inputs. 
In an analog circuit with input voltages $V_{\text{in}}^{+}$ and $V_{\text{in}}^{-}$, the ideal output voltage $V_{\text{out}}$ is proportional to the difference between the two inputs, as expressed by:  
\begin{equation}  
    V_{\text{out}} = A_{d} (V_{\text{in}}^{+} - V_{\text{in}}^{-}),  
\label{eq:diffamp}  
\end{equation}  
where $A_{d}$ represents the differential gain. 
% This formulation aligns with the principles of the differential transformer \cite{ye2025difftrans}.

\begin{figure}[!tb]
    \centering
    \includegraphics[width=0.828\linewidth]{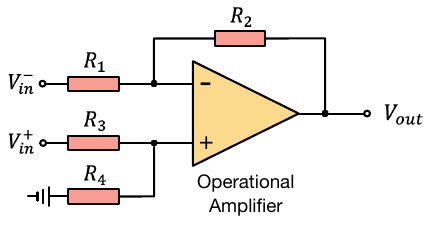} 
    \caption{The operational amplifier with two input voltages $V_{\text{in}}^{+}$ and $V_{\text{in}}^{-}$. 
    The CMRR $\mathcal{K}$ is controlled by resistances $R_1$, $R_2$, $R_3$, $R_4$.
    } 
    \label{fig:electronic-opamp}
\end{figure}

\minisection{Operational Amplifier}
In practical applications, the desired output often deviates from the predictions of \Cref{eq:diffamp}.
For instance, when $V_{\text{in}}^{+}$ and $V_{\text{in}}^{-}$ are equal, the output voltage does not necessarily become zero.
However, according to \Cref{eq:diffamp}, the output voltage should theoretically be zero in such cases. 
To address this discrepancy, as shown in \Cref{fig:electronic-opamp}, the OpAmp~\cite{sansen2007analog} provides a more accurate and stable output expression, including an additional term accounting for common-mode effects:
\begin{align}  
    V_{\text{out}} &= V_{\text{in}}^{+}\cdot(\frac{R_4}{R_3+R_4}\cdot\frac{R_1 + R_2}{R_1}) - V_{\text{in}}^{-}\cdot\frac{R_2}{R_1} \nonumber\\
    &= A_{d}(V_{\text{in}}^{+} - V_{\text{in}}^{-}) + \frac{A_{c}}{2}(V_{\text{in}}^{+} + V_{\text{in}}^{-}),
\label{eq:opampamp}  
\end{align}  
where $A_{c}$ is the common-mode gain of the amplifier. 
The common-mode rejection ratio (CMRR) is defined as the ratio of the differential gain to the common-mode gain:
\begin{equation}
    \mathcal{K} = \frac{A_{d}}{A_{c}}.
\label{eq:cmrr}
\end{equation}
Obviously, $A_{c} \rightarrow 0$ and $\mathcal{K} \rightarrow \infty$ for an ideal differential amplifier. 

\subsection{OpAmp Adaptation}  
\begin{figure}[!tb]
    \centering
    \includegraphics[width=0.928\linewidth]{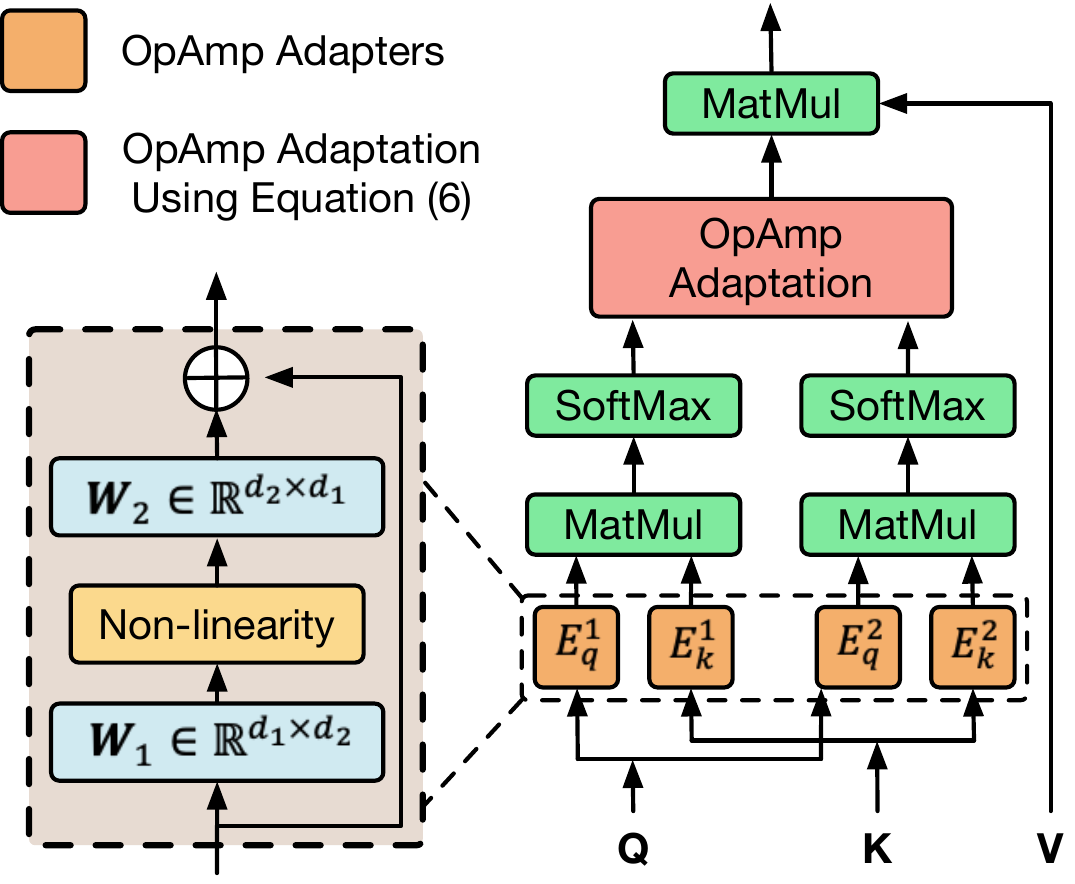} 
    \caption{Overview of the OpAmp adaptation using \Cref{eq:opampattn} with adapters.} 
    \label{fig:opamp}
\end{figure}

% In practical applications, the gains for the two input signals of an amplifier are not perfectly equal.
Inspired by the operational amplifier, we propose the OpAmp adaptation, which modifies the original attention mechanism into the OpAmp attention mechanism. 
Specifically, the operational amplifier is employed to denoise the input signals and produce a refined output in the analog circuit domain. 
Building on this concept, we design the OpAmp attention mechanism to denoise the attention matrices $\vec{M}$. 
As shown in \Cref{fig:opamp}, the original attention mechanism described in \Cref{eq:attn} is adapted using \Cref{eq:opampamp}:
\begin{equation}
    \vec{\bar{M}} = A_{d}(\vec{M}^{+} - \vec{M}^{-}) + \frac{A_{c}}{2}(\vec{M}^{+} + \vec{M}^{-}),
\label{eq:opampattn}   
\end{equation}
where $\vec{\bar{M}}$ is the denoised attention matrix via OpAmp adaptation, $\vec{M}^{+}$ and $\vec{M}^{-}$ are formulated through adapters, the detailed implementation of which will be elaborated in \Cref{sec:arch_des}.
As illustrated in \Cref{eq:opampattn}, we can adopt different $\mathcal{K}$ to adapt different scenarios using \Cref{eq:cmrr}.

Notably, the attention noise for LLMs after alignment is relatively small in noisy-context scenarios as shown in \Cref{fig:motivation}.
This suggests that attention denoising requires only a modest CMRR $\mathcal{K}$ instead of high CMRR values.
The experiment results presented in \Cref{sec:ab} further support our claim that excessively high CMRR values can lead to performance degradation. 

\subsection{Architecture Design}
\label{sec:arch_des}

Given an input feature $\vec{X} \in \mathbb{R}^{N \times d}$, the query feature $\vec{Q} \in \mathbb{R}^{N \times d}$ and the key feature $\vec{K} \in \mathbb{R}^{N \times d}$ are computed as follows:
\begin{equation}
    \vec{Q} = \vec{X}\vec{W}^q, \vec{K} = \vec{X}\vec{W}^k,
\end{equation}
where $\vec{W}^q, \vec{W}^k \in \mathbb{R}^{d \times d}$ represent pre-trained weights used for linear projection.
As outlined in \Cref{eq:opampattn}, the computation of \(\vec{M}^+\) and \(\vec{M}^-\) is required to implement the OpAmp attention mechanism.
A straightforward approach involves duplicating $\vec{W}^Q$ and $\vec{W}^K$ to compute two sets of query and key features, denoted as $\vec{Q}_1, \vec{K}_1$ and $\vec{Q}_2, \vec{K}_2$
Subsequently, $\vec{M}^+$ and $\vec{M}^-$ can be calculated independently using \Cref{eq:attn} as follows:
\begin{align}
    \vec{M}^+ &= \mathrm{Softmax}\left(\frac{\vec{Q}_1\vec{K}_1^{\top}}{\sqrt{d}}\right), \\
    \vec{M}^- &= \mathrm{Softmax}\left(\frac{\vec{Q}_2\vec{K}_2^{\top}}{\sqrt{d}}\right),
\end{align}
However, this method incurs substantial computational overhead, particularly given the large parameter scale of LLMs. 

Consequently, we introduce an effective and efficient implementation of OpAmp adaptation to address this inefficiency.
Specifically, we employ adapters to avoid redundant weight computations as shown in \Cref{fig:opamp}.
For a given input $\vec{X}$, the query and key features $\vec{Q}_1, \vec{K}_1$ and $\vec{Q}_2, \vec{K}_2$ can be computed as follows:
\begin{align}
    \vec{Q}_1 = E^1_q(\vec{X}\vec{W}^q), \vec{Q}_2 = E^2_q(\vec{X}\vec{W}^q), \\
    \vec{K}_1 = E^1_k(\vec{X}\vec{W}^k), \vec{K}_2 = E^2_k(\vec{X}\vec{W}^k),
\end{align}
where $E^{i}_{j}(\vec{x})$ represents the adapters for OpAmp adaptation, defined according to \Cref{eq:adapter} as:
\begin{equation}
    E^{i}_{j}(\vec{x}) = \phi(\vec{x}\vec{W}_1)\vec{W}_2 + \vec{x},
\label{eq:opampadapter}
\end{equation}
with $i \in \{1, 2\}$ and $j \in \{q, k\}$.
This architecture ensures effective OpAmp adaptation while minimizing computational overhead.
Finally, the output of OpAmp attention can be computed as:
\begin{equation}
    \mathrm{OpAmpAttn}(\vec{Q}, \vec{K}, \vec{V}) = \vec{\bar{M}}\vec{V}.
\end{equation}
\minisection{Zero Initialization}  
At the onset of training, we employ zero initialization to promote identity mapping. 
Specifically, $\vec{W}_2$ is initialized to zero to guarantee that $E^{i}_{j}(\vec{x}) = \vec{x}$. 
Furthermore, to prevent any disruption to the original $\vec{M}$ during the initial phase of training, we set $A_c = 1$ and regulate $\mathcal{K} = \frac{A_d}{A_c}$ by adjusting the values of $A_d$. 
As a result, at the initial stage, \Cref{eq:opampattn} reduces to:
\begin{align}
    \vec{\bar{M}} &= A_{d} \cdot (\vec{M} - \vec{M}) + \frac{A_{c}}{2} \cdot (\vec{M} + \vec{M}), \nonumber \\
    &= A_{d} \cdot 0 + \frac{A_{c}}{2} \cdot 2 \vec{M} = \vec{M},
\end{align}
which aligns with the standard attention mechanism outlined in \Cref{eq:attn}. 
This strategy ensures that the model initiates training with a well-established mechanism before incorporating more sophisticated modifications.
Moreover, other modules, such as the normalization and FFN layers, are replicated directly from the original transformer block to ensure structural coherence.
\section{Experiments}

\begin{table*}[t]
\centering
\renewcommand{\arraystretch}{1.0}
\resizebox{0.808\linewidth}{!}{
\footnotesize
\begin{tabularx}{0.878\linewidth}{X|cc|cc|cc}
\toprule
& \makecell{Qwen2.5 \\ OpAmp-72B} & \makecell{Llama3 \\ ChatQA2-70B} & \makecell{Qwen2.5 \\ 72B inst} & \makecell{Llama3.3 \\ 70B inst} & \makecell{DeepSeek \\ V3} & \makecell{GPT-4o \\ 0806} \\
\midrule
\makecell[X]{LooGLE (EM) \newline \cite{li2023loogle}}
& \textbf{66.3} & 59.1 & 64.9 & 63.0 & 63.4 & 62.7 \\ 
\midrule
\makecell[X]{NarrativeQA (EM) \newline \cite{kovcisky2018narrativeqa}}
& \textbf{61.7} & 59.8 & 60.2 & 61.5 & 60.5 & 61.5 \\  
\midrule
\makecell[X]{MultiHopRAG (EM) \newline \cite{tang2024multihoprag}} 
& \textbf{89.6} & 78.2 & 89.2 & 83.7 & 88.6 & 87.7\\ 
\midrule
\makecell[X]{HotpotQA (EM) \newline \cite{yang2018hotpotqa}}    
& \textbf{77.5} & 70.5 & 76.0 & 74.5 & 77.0 & \textbf{77.5}\\ 
\midrule
\makecell[X]{MuSiQue (EM) \newline \cite{trivedi2022musique}} 
& 48.0 & 39.0 & 44.0 & 47.5 & 52.5 & \textbf{53.0}\\ 
\midrule
\makecell[X]{CoQA (EM) \newline \cite{reddy2019coqa}}   
& \textbf{92.4} & 80.2 & 85.8 & 88.2 & 88.4 & 88.6\\  
\bottomrule
\end{tabularx}}
\caption{Performance of Qwen2.5-OpAmp-72B on various noisy context benchmarks. 
We present a detailed comparison of the Qwen2.5-OpAmp-72B with current SOTA open-source and commercial LLMs. 
We bold the highest scores among all models.}
\label{table:main_best}
\end{table*}

\begin{table*}[t]
\centering
\renewcommand{\arraystretch}{1.0}
\setlength\tabcolsep{9.6pt}
\resizebox{0.808\linewidth}{!}{
\footnotesize
\begin{tabularx}{0.862\linewidth}{X|cc|ccc}
\toprule
& \makecell{Llama3.1 \\ OpAmp-8B} & \makecell{Llama3 \\ ChatQA2-8B} & \makecell{Mistral \\ 7B inst-v0.3} & \makecell{Llama3.1 \\ 8B inst} & \makecell{Qwen2.5 \\ 7B inst} \\
\midrule
\makecell[X]{LooGLE (EM)}  
& \textbf{56.6} & 50.7 & 51.6 & 56.1 & 53.8 \\ 
\midrule
\makecell[X]{NarrativeQA (EM)}
& \textbf{57.4} & 53.1 & 44.7 & 55.9 & 47.7 \\ 
\midrule
\makecell[X]{MultiHopRAG (EM)}    
& \textbf{70.5} & 50.9 & 69.5 & 63.9 & 66.9 \\ 
\midrule
\makecell[X]{HotpotQA (EM)}   
& \textbf{61.0} & 56.5 & 58.0 & 58.5 & 59.5 \\
\midrule
\makecell[X]{MuSiQue (EM)} 
& \textbf{35.0} & 23.0 & 28.5 & 29.5 & 31.5 \\ 
\midrule
\makecell[X]{CoQA (EM)}   
& \textbf{85.4} & 78.2 & 70.6 & 82.2 & 84.2 \\ 
\bottomrule
\end{tabularx}}
\caption{Performance of Llama3.1-OpAmp-8B on various noisy context benchmarks. 
We present a detailed comparison of the Llama3.1-OpAmp-8B with various open-source LLMs with similar parameters. 
We bold the highest scores among all models.}
\label{table:main_small}
\end{table*}

\subsection{Training Settings}

\minisection{Training Data} 
We incorporate some noisy context data into the general supervised fine-tuning dataset to enhance LLMs' denoising capability in noisy context scenarios.
This training involved integrating three distinct datasets: LongCite-45k~\cite{zhang2024longcite}, Neural-Bridge-RAG~\cite{NeuralBridge2024ragdataset} and Tulu3-SFT-Mix~\cite{lambert2024tulu3}.
% The contexts in the first two datasets are divided into several chunks, which include golden documents and noisy documents, to better simulate noisy context scenarios.
% Data augmentation is also performed to expand the noisy context data.
After data processing, we get the \textbf{N}oisy \textbf{C}ontext \textbf{F}ine-\textbf{T}uning (NCFT) dataset for supervised fine-tuning. 
We provide more details of the NCFT dataset in \Cref{sec:training_datasets}.

\minisection{OpAmp Models}
We select two pre-trained models with different model sizes, Qwen2.5-72B~\cite{yang2024qwen2_5} and Llama3.1-8B~\cite{dubey2024llama3}, as our base models to train our OpAmp models using the NCFT dataset.
Moreover, we use the QLoRA technique to update the other parameters in the pre-trained models instead of full fine-tuning.
Please refer to \Cref{sec:training_details} for more details.

\subsection{Evalutaion Settings}

\minisection{Baselines} 
We compare our OpAmp models with existing powerful LLMs in our evaluation benchmark.
These LLMs include Llama3-ChatQA2-70B~\cite{xu2024chatqa2}, Qwen2.5-72B-inst~\cite{yang2024qwen2_5}, Llama3.3-70B-inst~\cite{dubey2024llama3}, DeepSeek-V3~\cite{liu2024deepseekv3}, GPT-4o-0806~\cite{hurst2024gpt4o}, Llama3-ChatQA2-8B~\cite{xu2024chatqa2}, Mistral-7B-inst-v0.3~\cite{jiang2023mistral}, Llama3.1-8B-inst~\cite{meta2024llama3_1} and Qwen2.5-7B-inst~\cite{yang2024qwen2_5}.

\minisection{Evalution Benchmarks} 
Our evaluation benchmarks are designed using a spectrum of well-known datasets and benchmarks including LongBench~\cite{bai2024longbench} and ChatQA~\cite{liu2024chatqa}.
After some selection and filtration, these benchmarks can be categorized as follows:
\begin{itemize}[itemsep=0pt,topsep=0pt,parsep=0pt]
    \item \textbf{Long-Context QA:} The evaluation encompasses partial match (PM), exact match (EM), and accuracy (Acc.) metrics for various long-context QA benchmarks, including NarrativeQA~\cite{kovcisky2018narrativeqa}, Qasper~\cite{dasigi2021qasper}, QuALITY~\cite{pang2021quality}, and LooGLE~\cite{li2023loogle}.
    \item \textbf{Multi-Hop QA:} Assessment of multi-hop reasoning performance on various benchmarks, including HotpotQA~\cite{yang2018hotpotqa}, MuSiQue~\cite{trivedi2022musique}, and MultiHopRAG~\cite{tang2024multihoprag}, using the EM metric.
    \item \textbf{Noisy-RAG QA:} PM and EM scores for RAG scenarios using CoQA~\cite{reddy2019coqa}, QuAC~\cite{choi2018quac}, and QReCC~\cite{anantha2020qrecc} benchmarks.
\end{itemize}
For a more detailed composition of the evaluation benchmark, please refer to \Cref{sec:evaluation_benchmarks}.

\subsection{Evaluation on Noisy-Context Benchmarks}

We perform various experiments on LLMs with different sizes to evaluate the capabilities of our OpAmp adaptation.
For LLMs with more than 70B parameters, we compare Qwen2.5-OpAmp-72B with Llama3-ChatQA2-70B, Qwen2.5-72B-inst, Llama3.3-70B-inst, DeepSeek-V3, and GPT-4o-0806.
For LLMs with around 7B parameters, we compare Llama3.1-OpAmp-8B with Llama3-ChatQA2-8B, Mistral-7B-inst-v0.3, Llama3.1-8B-inst, and Qwen2.5-7B-inst.
The noisy-context benchmarks cover a wide range of tasks.
For long-context scenarios, LooGLE and NarrativeQA are selected.
We utilize MultiHopRAG, HotpotQA, and MuSiQue for Multi-Hop reasoning evaluation and CoQA for noisy-RAG scenarios.
\Cref{table:main_best} and \Cref{table:main_small} demonstrate the superior performance of our OpAmp models compared to other existing powerful LLMs, underscoring the significant capabilities and effectiveness of the OpAmp adaptation in noisy context scenarios.
% regardless of the base model utilized.

\begin{table*}[tb!]
\centering
\setlength\tabcolsep{3.6pt}
\resizebox{0.948\linewidth}{!}{
\begin{tabular}{lc|c|cccc|ccc|ccc}
\toprule
\multirow{2}{*}{Method} & \multirow{2}{*}{$\mathcal{K}$} & \multirow{2}{*}{Avg.} & Qasper & LooGLE & NarrativeQA & QuALITY & MultiHopRAG & HotpotQA & MuSiQue & CoQA & QuAC & QReCC \\ 
&  &  & (PM) & (EM) & (EM) & (Acc.) & (EM) & (EM) & (EM) & (EM) & (PM) & (PM) \\ 
\midrule
QLoRA & - & 52.4 & 38.9 & 53.1 & 55.7 & 76.1 & 68.4 & 56.5 & 31.5 & 83.6 & 25.2 & 35.4 \\
\midrule
\multirow{4}{*}{\makecell{OpAmp \\ Adapter}}
& 1  & 54.1 (+1.7) & 40.8 & 56.0 & 56.4 & 79.2 & 68.5 & 57.5 & 32.5 & \textbf{85.8} & 26.1 & 38.3 \\
& 5  & 54.3 (+1.9) & 41.2 & 56.5 & 56.9 & 77.8 & 69.5 & \textbf{62.0} & 31.5 & 84.6 & 25.5 & 37.1 \\
& 10 & \textbf{55.4 (+3.0)} & \textbf{43.1} & \textbf{56.6} & \textbf{57.4} & 79.0 & 70.5 & 61.0 & \textbf{35.0} & 85.4 & \textbf{26.5} & \textbf{39.8} \\
& 20 & 54.4 (+2.0) & 41.5 & 55.4 & 56.4 & \textbf{79.3} & \textbf{71.4} & 59.0 & 33.0 & 84.0 & 26.2 & 37.0 \\
\bottomrule
\end{tabular}}
\caption{Ablation studies on various noisy context benchmarks using Llama3.1-8B-base as the base model.
We bold the highest scores for each benchmark.}
\label{table:ab_study_cmrr}
\end{table*}

\minisection{Long-Context Evaluation}
Long-context evaluation requires LLMs to disregard large volumes of context-related but question-irrelevant information within extensive texts, accurately identify the paragraphs relevant to the answer, and generate responses based on these pertinent segments. 
Our Qwen2.5-OpAmp-72B model achieves EM scores of up to 66.3\% on the LooGLE benchmark with a maximum context length of 32K tokens and 61.7\% on the NarrativeQA benchmark with a maximum context length of 64K tokens. 
Similarly, our Llama3.1-OpAmp-8B model attains the highest EM score of 56.6\% on the LooGLE benchmark and leads with a score of 57.4\% on the NarrativeQA benchmark. 
These experiment results underscore the robust capability of our OpAmp models to filter out context-related noise and accurately locate answers within long contexts. 
Furthermore, they demonstrate the strong generalization ability of our approach across different model sizes.

\begin{table}[tb!]
\centering
\setlength\tabcolsep{2.4pt}
\resizebox{0.988\linewidth}{!}{
\begin{tabular}{ll|ccc|ccc|ccc}
\toprule
&  & \multicolumn{3}{c|}{CoQA (EM)} & \multicolumn{3}{c|}{QuAC (PM)} & \multicolumn{3}{c}{QReCC (PM)} \\
\midrule
% \multirow{2}{*}{Model} & \multirow{2}{*}{CMRR}
\multicolumn{2}{c|}{Noise Ratio} & 0.0 & 0.8 & 0.9 & 0.0 & 0.8 & 0.9 & 0.0 & 0.8 & 0.9 \\ 
\midrule
\multicolumn{2}{c|}{QLoRA} & 89.8 & 85.4 & 83.6 & 27.5 & 26.1 & 25.2 & 36.5 & 36.4 & 35.4 \\
\midrule
\multirow{4}{*}{\makecell{OpAmp \\ Adapter}}
& \multirow{4}{*}{$\mathcal{K} = \left\{
    \begin{tabular}{c}
    1 \\ 
    5 \\ 
    10 \\
    20
    \end{tabular}
\right.$}  
& 90.4 & 85.6 & \textbf{85.8} & 28.5 & 26.2 & 26.1 & 39.4 & 39.1 & 38.3 \\
& & 90.0 & 85.6 & 84.6 & 27.5 & 26.7 & 25.5 & 38.2 & 37.3 & 37.1 \\
& & 91.2 & \textbf{88.0} & 85.4 & 28.5 & 26.5 & \textbf{26.5} & \textbf{40.8} & \textbf{39.8} & \textbf{39.8} \\
& & \textbf{91.8} & 86.6 & 84.0 & \textbf{28.6} & \textbf{28.0} & 26.2 & 38.5 & 38.1 & 37.0 \\
\bottomrule
\end{tabular}}
\caption{Ablation studies on various benchmarks with different noise ratios using Llama3.1-8B-base as the base model.
We bold the highest scores.}
\label{table:ab_study_noise}
\end{table}

\minisection{Multi-Hop Evaluation}
Multi-hop evaluation is designed to assess the capability of LLMs to extract and synthesize relevant information from multiple documents for reasoning. 
This task requires LLMs to filter out irrelevant or noncritical documents to minimize interference during the reasoning process.
Our Qwen2.5-OpAmp-72B model demonstrates strong performance on multi-hop reasoning tasks, achieving high scores of 89.6\% on MultiHopRAG and 77.5\% on HotpotQA, with notable advantages over competing models. 
Although it performs slightly weaker than top-performing LLMs on the MuSiQue benchmark, its EM score of 48.0\% remains competitive for multi-hop reasoning tasks. 
Additionally, our Llama3.1-OpAmp-8B model also excels in multi-hop reasoning benchmarks, achieving top scores of 70.5\% on MultiHopRAG, 61.0\% on HotpotQA, and 35.0\% on MuSiQue, consistently surpassing other models. 
These results highlight the superior ability of our OpAmp models to handle complex, multi-step reasoning tasks across various benchmarks, underscoring its effectiveness in enhancing reasoning capabilities.

\minisection{Noisy-RAG Evaluation}
For the currently most widely adopted RAG technology, we conduct the noisy-RAG evaluation to assess the ability of LLMs to filter out irrelevant documents and accurately identify the document containing the correct answer in real-world RAG scenarios.
Our Qwen2.5-OpAmp-72B model achieves a top score of 92.4\% on the CoQA benchmark, surpassing the second-closest LLM, DeepSeek-V3, by a significant margin of 4\%. 
Our Llama3.1-OpAmp-8B model also attains a leading score of 85.4\% on the CoQA benchmark, outperforming Qwen2.5-7B-inst by 1.2\%. 
These experimental results highlight the superior performance of our OpAmp models in identifying correct answers within real-world RAG scenarios, exhibiting robust resistance to interference and noise.

\subsection{Ablation Studies}
\label{sec:ab}

To further investigate the contribution of $\mathcal{K}$, we perform a series of ablation studies. 
Additionally, we compare our OpAmp approach with the QLoRA technique. 
In brief, we denote the OpAmp adapter as adapters implemented for our OpAmp adaptation.
To ensure fair comparisons in these ablation studies, both OpAmp and QLoRA models are fine-tuned using the same dataset, NCFT.

\minisection{CMRR}
\Cref{table:ab_study_cmrr} presents a comparative analysis of QLoRA and the OpAmp adapter for enhancing the Llama3.1-8B-base model across various noisy context benchmarks. 
The OpAmp adapter demonstrates consistent superiority over QLoRA across all evaluated benchmarks. 
Specifically, QLoRA achieves an average score of 52.4\%, whereas the OpAmp adapter significantly enhances performance, with the best results observed at $\mathcal{K}=10$, yielding an average score of 55.4\%. 
% Notably, the OpAmp adapter with $\mathcal{K}=10$ attains the highest average score, highlighting its efficacy in managing noisy-context tasks.
When examining the impact of different values of $\mathcal{K}$, $\mathcal{K}=10$ emerges as the optimal configuration across multiple benchmarks. 
Larger value ($\mathcal{K}=20$) exhibits diminishing returns, while smaller values ($\mathcal{K}=1, 5$) perform adequately but are marginally less competitive. 
This suggests our statement that attention denoising requires only a modest $\mathcal{K}$ instead of the $\mathcal{K} \rightarrow \infty$ used in the differential transformer architecture~\cite{ye2025difftrans}.

\begin{table}[tb!]
\centering
\setlength\tabcolsep{2.0pt}
\resizebox{0.988\linewidth}{!}{
\begin{tabular}{lc|ccc|c}
\toprule
\multirow{3}{*}{Method} & \multirow{3}{*}{$\mathcal{K}$} & \multicolumn{4}{c}{FaithEval} \\
\cmidrule{3-6}
& & Inconsistent & Unanswerable & Counterfactual & \multirow{2}{*}{Avg.}\\ 
& & (EM) & (EM) & (EM) & \\ 
\midrule
QLoRA & - & 24.1 & 46.1 & 71.6 & 47.3\\
\midrule
\multirow{4}{*}{\makecell{OpAmp \\ Adapter}}
& 1 & \textbf{45.5} & 53.1 & \textbf{76.3} & \textbf{58.3 (+11.0)} \\
& 5 & 42.1 & \textbf{53.7} & 75.9 & 57.2 (+9.90) \\
& 10 & 45.3 & 53.0 & 75.1 & 57.8 (+10.5) \\
& 20 & 22.3 & 58.8 & 73.8 & 51.6 (+4.30) \\
\bottomrule
\end{tabular}}
\caption{Ablation studies on FaithEval using Llama3.1-8B-base as the base model.
We bold the highest scores.}
\label{table:ab_study_faitheval}
\end{table}

\minisection{Noise Ratio}
The ablation study detailed in \Cref{table:ab_study_noise} assesses the performance of QLoRA and the OpAmp adapter across varying noise ratios (0.0, 0.8, 0.9) on noisy-RAG benchmarks, including CoQA, QuAC, and QReCC. 
The noise ratio is simulated by introducing noise documents into the original golden document, replicating increasingly challenging real-world RAG scenarios. 
As expected, performance across all methods generally degrades with increasing noise ratios, reflecting the growing difficulty of extracting relevant information from cluttered contexts.
QLoRA exhibits a steady decline in performance as noise levels increase. 
For instance, its score on CoQA drops from 89.8\% at a noise ratio of 0.0 to 83.6\% at 0.9. 
In contrast, the OpAmp adapter demonstrates greater robustness, particularly when configured with $\mathcal{K}=10$. 
Moreover, higher values of $\mathcal{K}$ occasionally underperform compared to $\mathcal{K}=10$, indicating that excessive attention denoise may compromise the capability.
Overall, the OpAmp adapter consistently outperforms QLoRA across all noise levels, with $\mathcal{K}=10$ emerging as the optimal configuration for balancing robustness and performance under noisy conditions. 
This underscores the effectiveness of our method in handling challenging RAG scenarios.

\minisection{Hallucination}
As shown in \Cref{table:ab_study_faitheval}, the ablation study on FaithEval~\cite{ming2024faitheval} demonstrates that OpAmp not only enhances robustness to noisy contexts but also reduces hallucinations as a valuable secondary benefit. 
While QLoRA achieves an average score of 47.3\%, OpAmp attains much higher averages, with the best results with $\mathcal{K}=1$ (58.3\%), indicating consistent improvements. 
Notably, $\mathcal{K}=1, 5, 10$ exhibit similar performance levels, suggesting that moderate values of $\mathcal{K}$ effectively balance denoising and model stability while mitigating hallucinations. 
However, performance declines significantly (51.6\%) when $\mathcal{K}=20$.
The degradation demonstrates an excessive attention-denoising process caused by excessive CMRR, which impairs the model's ability to avoid hallucination.
This analysis underscores that the optimal performance is achieved with moderate $\mathcal{K}$ values, highlighting the importance of balancing denoising intensity with model adaptability.

\subsection{Visualization of Attention}

\begin{figure}[!tb]
    \centering
    \includegraphics[width=0.968\linewidth]{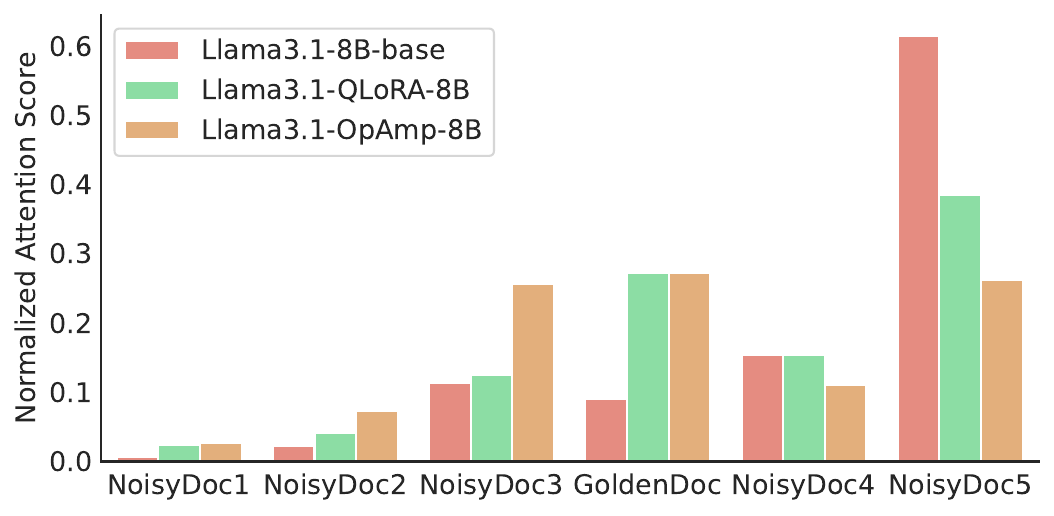} 
    \caption{Normalized attention score. Our OpAmp model demonstrates significant attention denoise capability compared to the base model and QLoRA model.} 
    \label{fig:abstudy}
\end{figure}

\begin{figure}[!tb]
    \centering
    \includegraphics[width=0.968\linewidth]{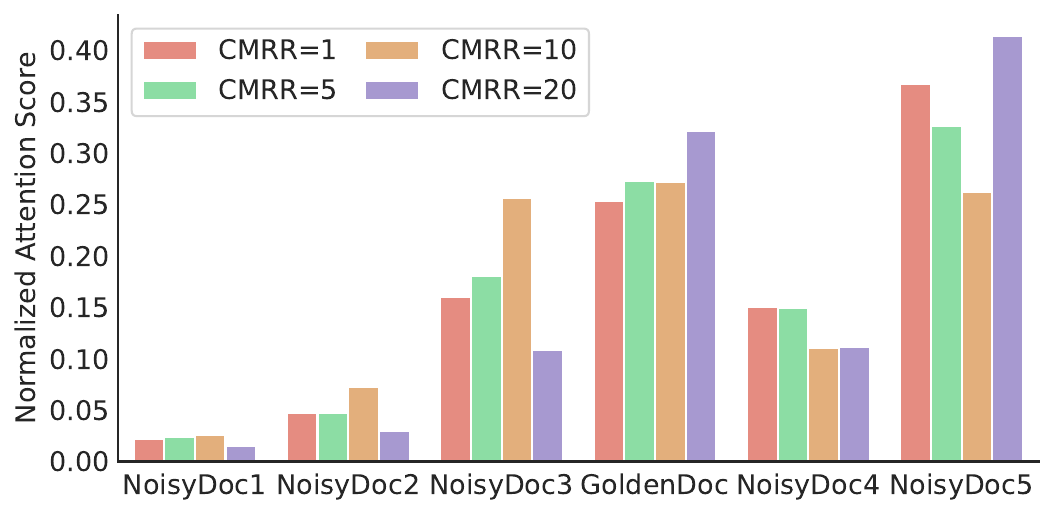} 
    \caption{Normalized attention score with different values of $\mathcal{K}$ utilizing for OpAmp adaptation.} 
    \label{fig:cmrr}
\end{figure}

To provide deeper insights into the OpAmp mechanism, we perform some visualizations of $\vec{\bar{M}}$. 
As previously mentioned, transformer-based architectures tend to allocate disproportionate attention to irrelevant or later-positioned documents. 
In contrast, OpAmp can enhance LLMs' focus on the most relevant documents. 
We employ normalized attention scores based on Llama3.1-8B to trace the OpAmp mechanism in a noisy context to investigate this behavior.
As shown in \Cref{fig:abstudy}, Llama3.1-8B-base becomes completely lost in the noisy context, with its attention distribution across documents generally increasing sequentially from low to high.
Llama3.1-QLoRA-8B model performs relatively better, with a slight increase in attention to the golden document.
However, the limitation of a forced backward shift in attention still exists. 
In contrast, our Llama3.1-OpAmp-8B uniquely allocates the most attention to the golden document among all documents. 
This mechanism is a key factor contributing to the strong performance of our OpAmp model in noisy context scenarios.
Meanwhile, we also investigate the mechanism across different CMRR values.
As illustrated in \Cref{fig:cmrr}, only when $\mathcal{K}=10$ does the OpAmp model allocate the highest level of attention to the golden document, surpassing the other CMRR values and indirectly confirming that a moderate CMRR value is crucial for maximizing the effectiveness of the OpAmp mechanism instead of $\mathcal{K}\rightarrow \infty$ utilized in differential transformer~\cite{ye2025difftrans}.
\section{Related Works}

\subsection{Question Answering with Noisy Contexts}
The internal knowledge of LLMs often fails to meet diverse application needs~\cite{he2022retrieval, Ji2023Survey}, driving research into integrating external knowledge. 
Among the proposed solutions~\cite{guu2020realm, beltagy2020longformer, wang2024knowledgeediting}, RAG~\cite{borgeaud2022retrieving, ren2024knowledgeboundary} and long-context modeling techniques~\cite{press2022trainshorttestlong, chen2023extendingcontextwindow} have emerged as two prominent strategies for incorporating external knowledge stored in long-text formats. 
However, recent studies~\cite{Shi2023distracted, liu2024lost, lv2024coarse, ye2025difftrans} have identified a significant challenge. 
Specifically, as the number of retrieved documents grows and the length of input contexts expands, the model is increasingly exposed to noise, which is often the non-critical information unrelated to the query.
This noisy-context scenario significantly degrades the performance of LLMs on QA tasks~\cite{chen2023benchmark}. 
Consequently, we propose the OpAmp adaptation with adapters, a plug-and-play solution that minimizes noisy context impact with low computation costs, enhancing the performance in such scenarios.

\subsection{Parameter Efficient Fine-Tuning}
Traditionally, full fine-tuning is the predominant approach for fine-tuning pre-trained models, including LLMs. 
However, this method entails substantial computational costs, particularly regarding time consumption and GPU memory usage. 
To address these challenges, a variety of PEFT methods have been developed \cite{houlsby2019adapter, hu2021lora, dettmers2024qlora, wu2024pesc, li2021prefixtuning, lester2021prompttuning}, enabling efficient fine-tuning without compromising performance compared to full fine-tuning. 
PEFT focuses on training a limited subset of parameters within the existing model or newly inserted modules.
Adapter-based methods \cite{houlsby2019adapter, hu2021lora, dettmers2024qlora, wu2024pesc} insert learnable modules into Transformer blocks, which contain a small number of parameters. 
These adapters are fine-tuned instead of the original model weights. 
Among these methods, QLoRA \cite{dettmers2024qlora} has gained significant attention for its efficiency in fine-tuning LLMs while maintaining performance comparable to full fine-tuning.
Another emerging trend in PEFT is prefix-tuning \cite{lester2021prompttuning, li2021prefixtuning}, which involves adding learnable token vectors to the input sequence. 
% However, prefix-tuning often exhibits inferior performance compared to adapter-based approaches. 
In this study, we introduce adapters to perform OpAmp adaptation. 
Specifically, adapters reformulate the computation of the original attention mechanism into the OpAmp attention mechanism.
% as shown in \Cref{eq:opampattn}.

\subsection{Adaptation of Pre-trained Models}
Recent studies~\cite{chen2015net2net, lin2021m610t, komatsuzaki2022sparseupcycle, wu2024pesc} have focused on improving training efficiency by leveraging pre-trained model weights for a warm start, thus accelerating convergence and minimizing training costs.
\citet{komatsuzaki2022sparseupcycle} and \citet{wu2024pesc} introduce methods to initialize sparse MoE models using weights from a pre-trained dense model.
These approaches significantly reduce the required training resources.
In this paper, we train our OpAmp models with OpAmp attention blocks using weights from pre-trained LLMs.
\section{Conclusion}
Inspired by the operational amplifiers, we introduce the OpAmp adaptation implemented with adapters in this study. 
By integrating this adapter into pre-trained Transformer blocks, our approach enhances the model's ability to focus on the most relevant context without expensive full-scale training from scratch. 
We implement our OpAmp models and other baselines with our noisy-context fine-tuning dataset, NCFT, for fair comparisons.
The OpAmp adaptation demonstrates significant performance gains across LLMs of varying model sizes. 
Extensive empirical evaluations are conducted on extensive noisy-context benchmarks. 
The results indicate that our Qwen2.5-OpAmp-72B model, fine-tuned with our OpAmp adaptation, outperforms current SOTA LLMs,
including DeepSeek-V3~\cite{liu2024deepseekv3} and GPT-4o~\cite{hurst2024gpt4o}.

%\newpage
\clearpage
\section*{Limitation}
The OpAmp adaptation with adapters introduces a marginally higher number of parameters compared to the standard PEFT training process with QLoRA. 
Consequently, the supervised fine-tuning process for our OpAmp models demands slightly greater GPU memory allocation and computational time. 
Additionally, our OpAmp models incur a minor latency during inference when compared to the original pre-trained LLMs.
\bibliography{ref/Top,ref/reference}

\clearpage

\appendix
\begin{table}[!tb]
\centering
\footnotesize
\resizebox{0.988\linewidth}{!}{
\begin{tabular}{cccccc}
\toprule 
 lr & epoch & LoRA $r$ & LoRA $\alpha$ & Adapter Dim \\
\midrule
$2 \times 10^{-4}$   &1    &64    &16  &512 \\
\bottomrule
\end{tabular}
}
\caption{Hyperparameters of supervised fine-tuning.}
\label{table:hyperparameters}
\end{table}

\begin{table}[!tb]
\centering
\setlength\tabcolsep{3.2pt}
\footnotesize
\resizebox{0.988\linewidth}{!}{
\begin{tabular}{lccc}
\toprule 
 & LongCite-45k & Neural-Bridge-RAG & Tulu3-SFT-Mix \\
\midrule
NCFT    &30k    &20k    &450k \\
\bottomrule
\end{tabular}
}
\caption{The proportion of LongCite-45k, Neural-Bridge-RAG and Tulu3-SFT-Mix in the NCFT dataset.}
\label{table:proportion}
\end{table}

\section{Implementation Details}
\label{sec:training_details}
The training process entailed using a constant learning rate schedule with a warm-up ratio of 0.03, and the paged AdamW \cite{dettmers2024qlora, loshchilov2017adamw} optimizer with a learning rate of $2 \times 10^{-4}$, no weight decay, a batch size of 128, and a sequence length of 8192 tokens. 
The models underwent instruction tuning for one epoch on 16 A100 GPUs, each with 80G memory.

Moreover, we employed the QLoRA \cite{dettmers2024qlora} technique for efficient fine-tuning. 
As for the QLoRA configuration, we use a 4-bit quantization scheme for our experiments, which significantly reduces memory usage while preserving model performance.
We show the hyperparameters for supervised fine-tuning in \Cref{table:hyperparameters}.

\section{Training Datasets}
\label{sec:training_datasets}

As shown in Table~\ref{table:proportion}, we shows the proportion of LongCite-45k~\cite{zhang2024longcite}, Neural-Bridge-RAG~\cite{NeuralBridge2024ragdataset} and Tulu3-SFT-Mix~\cite{lambert2024tulu3} in the NCFT dataset.

Considering the original format and quantity of LongCite-45k and Neural-Bridge-RAG, we perform data processing to simulate the noisy context scenarios. 
Firstly, we filter the Chinese corpus and divide the context into several chunks.
Then we preserve the chunks with golden documents and introduce relevant or irrelevant chunks as noise.
Finally, we filter low-quality corpora (too long or too short).
We obtained our supervised fine-tuning dataset after data processing
which encompasses a wide range of topics, and the noise ratio in the dataset ranges from 0 to 1, aiming to cover a variety of real-world situations and use cases.

\section{Evaluation Benchmarks}
\label{sec:evaluation_benchmarks}

We show the details of the noisy-context evaluation benchmark in \Cref{table:noisy-context benchmark}. 
Qasper, HotpotQA, and MuSiQue are directly derived from the LongBench~\cite{bai2024longbench}.
In contrast, CoQA, QuAC, and QReCC are QA datasets selected from ChatQA~\cite{liu2024chatqa} and have been noise-augmented in a manner consistent with \Cref{sec:training_datasets} to align with the noisy-RAG format. 
For the QuALITY dataset, we retain only the subset labeled as ``hard''.
Similarly, for the NarrativeQA, Loogle, and MultiHopRAG datasets, we apply filters based on context length and response quality to further enhance the benchmark's ability to differentiate between models.

\begin{table}[!tb]
\centering
\setlength\tabcolsep{3.2pt}
\footnotesize
\resizebox{0.988\linewidth}{!}{
\begin{tabular}{lcccc}
\toprule 
Benchmark & Source & Max Length & Metric &\# Data \\
\midrule
\multicolumn{5}{l}{\textit{Long-Context QA}} \\ 
\midrule
NarrativeQA & Literature, Film & 64K & EM& 1009 \\ 
Qasper  & Science & 8K  & PM & 200 \\ 
QuALITY & Literature & 8K  & Acc. & 1065 \\ 
LooGLE  & Science & 32K  & EM & 1427 \\
\midrule
\multicolumn{5}{l}{\textit{Multi-Hop QA}} \\
\midrule
HotpotQA & Wikipedia & 16K & EM & 200 \\  
MuSiQue & Wikipedia & 16K & EM & 200 \\ 
MultiHopRAG & News & 8K & EM & 2255 \\
\midrule
\multicolumn{5}{l}{\textit{Noisy-RAG QA}} \\
\midrule
CoQA & Multi-field & 4K & EM & 500 \\ 
QuAC & Wikipedia   & 4K & PM & 996 \\ 
QReCC & Multi-field & 4K & PM & 643 \\
\bottomrule
\end{tabular}
}
\caption{An overview of the dataset statistics for the noisy-context benchmark. The `Source' column indicates the origin of the context.}
\label{table:noisy-context benchmark}
\end{table}

% \section{Case Studies}

\end{document}